\documentclass[letterpaper]{article} % DO NOT CHANGE THIS
\usepackage{aaai2026}  % DO NOT CHANGE THIS
\usepackage{times}  % DO NOT CHANGE THIS
\usepackage{helvet}  % DO NOT CHANGE THIS
\usepackage{courier}  % DO NOT CHANGE THIS
\usepackage[hyphens]{url}  % DO NOT CHANGE THIS
\usepackage{graphicx} % DO NOT CHANGE THIS
\usepackage{natbib}  % DO NOT CHANGE THIS AND DO NOT ADD ANY OPTIONS TO IT
\usepackage{caption} % DO NOT CHANGE THIS AND DO NOT ADD ANY OPTIONS TO IT
\usepackage{algorithm}
\usepackage{algorithmic}
\usepackage{newfloat}
\usepackage{listings}
\DeclareCaptionStyle{ruled}{labelfont=normalfont,labelsep=colon,strut=off} % DO NOT CHANGE THIS
\floatstyle{ruled}
\newfloat{listing}{tb}{lst}{}
\floatname{listing}{Listing}
\usepackage{array}
\usepackage{multirow}
\usepackage{amsmath}
\usepackage{tabularx}

\usepackage{tabularx}
\usepackage{booktabs}
\usepackage{amssymb}

\usepackage{booktabs}
\usepackage{multirow}
\usepackage{makecell}
\usepackage{graphicx}
\usepackage[table]{xcolor}
\usepackage{dcolumn}

\title{Evaluating LLMs for Police Decision-Making: \\A Framework Based on Police Action Scenarios}

\author{
Sangyub Lee\textsuperscript{\rm 1, \rm 2}, % A1: 소속 1, 2
Heedou Kim\textsuperscript{\rm 1, \rm 3},
Hyeoncheol Kim\textsuperscript{\rm 1}\protect\thanks{Corresponding Author} % A3: 소속 1 + 교신 저자 각주 유지
}
\affiliations {
\textsuperscript{\rm 1}Department of Computer Science and Engineering, Korea University, Seoul, Republic of Korea\\
\textsuperscript{\rm 2}Korea National Police University, Asan, Republic of Korea\\
\textsuperscript{\rm 3}Police Science Institute, Asan, Republic of Korea\\
\{yubii2, heedou123, harrykim\}@korea.ac.kr
}

\begin{document}

\maketitle

\begin{abstract}
The use of Large Language Models (LLMs) in police operations is growing, yet an evaluation framework tailored to police operations remains absent. While LLM's responses may not always be legally “incorrect”, their unverified use still can lead to severe issues such as unlawful arrests and improper evidence collection. To address this, we propose PAS (Police Action Scenarios), a systematic framework covering the entire evaluation process. Applying this framework, we constructed a novel QA dataset from over 8,000 official documents and established key metrics validated through statistical analysis with police expert judgements. Experimental results show that commercial LLMs struggle with our new police-related tasks, particularly in providing fact-based recommendations. This study highlights the necessity of an expandable evaluation framework to ensure reliable AI-driven police operations. We release our data and prompt template.
\end{abstract}

\begin{links}
    \link{Data \& Code}{https://github.com/Heedou/PASFramework}
\end{links}

\section{Introduction}

Recently, police officers face significant operational challenges. Given the high workload caused by frequent overtime, crime scene exposure, and a substantial number of cases, officers often experience significant stress, which increases the risk of errors in police operations or delays in case processing \cite{Stotland:health,Tan:workload,Vila:LongWork}. To address these challenges, the integration of Large Language Models (LLMs) as auxiliary tools has become increasingly common, presenting the potential to reduce time and human resource consumption. For example, police officers can now leverage state-of-the-art LLMs for tasks such as traffic accident analysis, automated police report generation, automatic phishing detection, and criminal investigations \cite{heedou,Sarzaeim:Smart,Tong:CPSD,Halford:policethreats,Adams:policereport,Jamal:phishing, kim2022named} This integration fosters expectations that, in the long run, LLMs will contribute to a safer public security service \cite{heedou,Sarzaeim:Smart}.

% \begin{figure}[H] % [H]를 사용하여 현재 위치에 고정 - float 금지 패키지
\begin{figure}[t] 
    \centering
    \scalebox{0.7}{
        \includegraphics[width=0.6\textwidth]{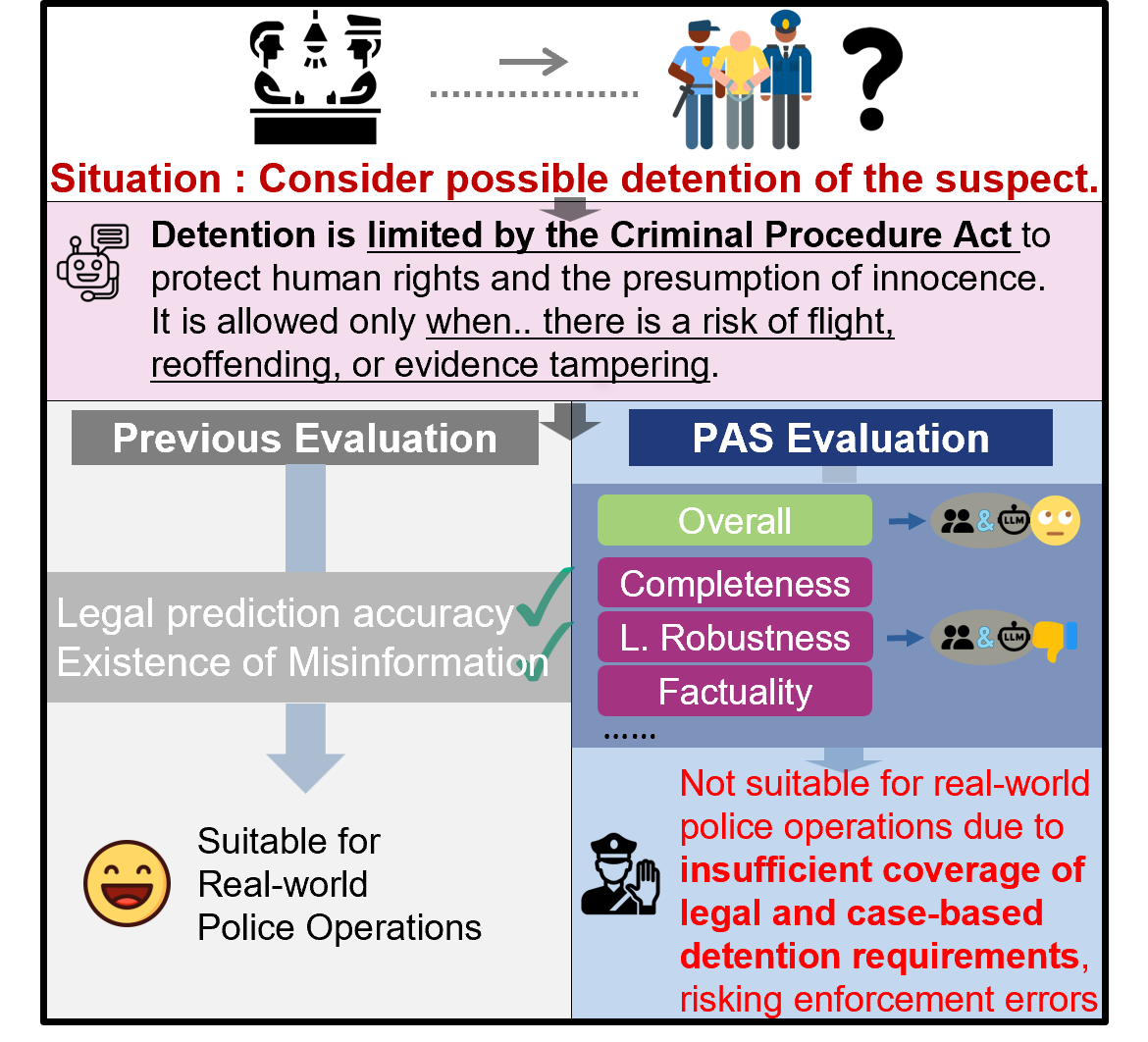}
    }
    \caption{
        LLMs are increasingly used to support police. However, existing evaluation focus only on information accuracy~\cite{heedou, Fei:Lawbench,hwang2022multi}, which risks indiscriminate use in real-world scenarios by overlooking key police-specific considerations. Our framework offers guidance for improving LLM from a policing perspective.
    }
    \label{fig:fig1}
\end{figure}

Yet, rigorous validation is essential for applying LLMs in specialized domains. Recent studies have conducted both quantitative and qualitative evaluations of LLM performance across various fields, including law and healthcare \cite{Liu:Healthcare,Liu:Medbench,Fei:Lawbench,hwang2022multi}. While these models sometimes achieve high accuracy, practical limitations remain. For example, in the legal domain, LLMs have successfully passed the bar exam \cite{katz2024gpt} and demonstrated high accuracy in Korean legal article prediction \cite{hwang2022multi}. However, in medical contexts, they achieved only a 65.6\% satisfaction rate for emergency treatment guidelines, highlighting their limited applicability in urgent situations \cite{birkun2023large}.

Similarly, the use of LLMs in policing must be carefully evaluated for real-world deployment, given the risks of biased, incomplete, or incorrect information, and overconfident responses. These concerns are not merely theoretical. Figure~\ref{fig:fig1} shows the risks of trusting LLMs in policing based solely on traditional evaluations. For example, LLM-based dispatch suggestions have shown regional and racial bias~\cite{Jain:cctv}. Police work involves critical tasks such as crime prevention and public safety~\cite{Policelaw,sikang}, which demand strict compliance with laws and procedures~\cite{roberts2012law}. As a single misstep can have serious consequences, LLM integration must proceed with caution and robust validation.

Despite this necessity, no comprehensive studies have yet assessed the suitability and associated risks of using LLMs in police activities. As shown in Figure~\ref{fig:fig1}, which highlights a case where traditional evaluation methods were applied to actual LLM outputs, serious issues may arise in practice when such models are used without proper evaluation from a policing perspective. Although LLM-generated responses may not be considered entirely incorrect in legal or informational terms, relying on them without verification could lead to legal and ethical problems, such as unlawful arrests or improper evidence collection.

To address these critical issues, this study aims to establish a framework for assessing the appropriateness of LLMs in police operations and to evaluate whether widely used LLMs meet the necessary standards. Ultimately, the goal is to determine the direction for developing future police-specific LLMs. Accordingly, this study defines the following two key research questions:
\begin{itemize}
    \item \textbf{RQ1}: What is an appropriate evaluation framework for LLMs in police operations that integrates domain-specific metrics and evaluation datasets?
    \item \textbf{RQ2}: To what extent can LLMs effectively answer inquiries about police work?
\end{itemize}

Based on the above research questions, this study identifies two major limitations in existing evaluations of LLMs for police work: (1) the lack of flexible scenario design methods and corresponding evaluation datasets tailored to real-world police response situations, and (2) the absence of a comprehensive metric framework for assessing LLM performance on police-specific tasks.

To address these gaps, we propose a specialized evaluation framework, \textbf{PAS} (Police Action Scenarios). \textbf{PAS} consists of five key stages: defining real-world police action scenarios, constructing expert reference answers for each scenario, generating LLM responses based on the scenarios, extracting core evaluation metrics, and interpreting results using LLM-based judges. The framework is designed to be flexibly adapted across diverse missions and operational contexts within the policing domain.

To validate the practical utility of \textbf{PAS}, we implemented it in the context of a \textit{Police Readiness through Operational Reasoning} scenario. We curated and refined over 8,000 official Korean police documents and constructed an evaluation dataset to benchmark the performance of commercial LLMs. Through the experiment, we identified five core evaluation metrics that are closely aligned with key indicators of real-world police performance. The results revealed that LLM-generated responses significantly underperformed expectations. These findings provide actionable insights for improving the future applicability of LLMs in police operations.

The contributions of our work are articulated as follows:

\begin{itemize}
    \item We propose \textbf{PAS}, a scalable evaluation framework tailored to police scenarios, covering various real-world policing scenario design, expert references, LLM responses, metric extraction, and judgment.
    \item We construct a curated corpus of 8,000+ Korean police documents and build an evaluation dataset for real-world LLM assessment. Field police experts participated in validating core evaluation metrics, highlighting the need for expert-in-the-loop supervision in developing reliable evaluation frameworks across diverse policing contexts.
    \item Our evaluation reveals that commercial LLMs underperform on key police metrics, highlighting the gap between general capabilities and domain-specific requirements.
\end{itemize}

\section{Related Works}

\begin{figure*}[h]
    \centering
    \includegraphics[width=0.9\textwidth]{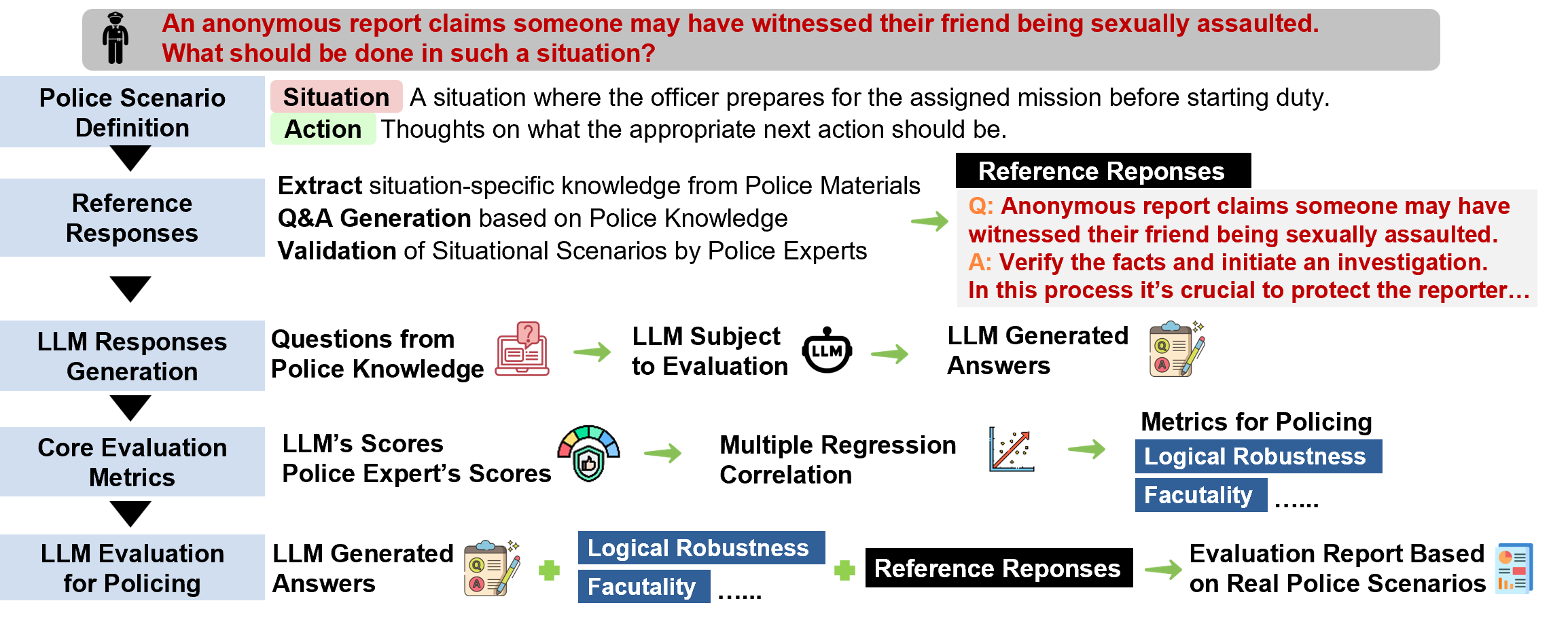}
	\caption{An overview of the \textbf{PAS}. Unlike previous benchmark studies that primarily focused on accuracy in legal matching or crime classification~\cite{heedou, kwon2024ai,hwang2022multi, baek2021smart}, this study adopts a Police Action Scenario-based framework to evaluate LLMs. By simulating real-world police situations, the framework generates both LLM responses and expert reference answers. Furthermore, it designs domain-specific evaluation metrics to comprehensively assess the LLM’s situational applicability in policing contexts.
	}
	\label{fig:fig2}
\end{figure*}

\subsection{LLM Evaluation Methods}

The evaluation of Large Language Models (LLMs) has evolved through two main approaches: multiple-choice QA and open-ended QA.\cite{myrzakhan2024open}

Multiple-choice QA has demonstrated remarkable effectiveness in assessing professional knowledge, with LLMs showing impressive performance across various professional certification exams including bar examination, medical licensing tests, and CPA evaluations\cite{katz2024gpt,kung2023performance,bommarito2023gpt}.

In parallel, open-ended QA assessment have undergone significant development. While traditional metrics like BLEU and ROUGE based on text similarity remain valid in certain contexts, they have been complemented by more sophisticated approaches that encompass multiple dimensions such as consistency, logical reasoning, and factual accuracy\cite{kamalloo2023evaluating}. While such multi-dimensional analysis traditionally relied on human evaluators, recent studies have shown that LLMs can effectively serve as automated judges, providing a scalable solution while maintaining quality standards \cite{chiang2023closer}.

\subsection{Challenges in Domain-Specific LLM Evaluation}

The application of these evaluation methodologies in professional domains has revealed significant limitations. Multiple-choice QA, despite its effectiveness in standardized tests, fails to capture practical competence in real-world scenarios\cite{li2024can}. In healthcare, LLMs achieved only 65.6\% satisfactory responses when evaluating emergency treatment guidelines \cite{birkun2023large}. Similarly, in the legal domain, while LLMs demonstrate impressive results in structured evaluations\cite{hwang2022multi}, their performance significantly drops when faced with real-world police scenarios. Additional issues include potential biases in their decision-making processes\cite{Jain:cctv}.
For open-ended QA, traditional metrics like BLEU and ROUGE are insufficient for evaluating domain-specific responses. Even with reference answers available, high BLEU or ROUGE scores may not indicate better responses - in fact, responses with high textual similarity might contradict the intended meaning or provide inappropriate guidance in professional contexts\cite{xu2024reasoning}. While multi-dimensional evaluation approaches offer a potential solution, the lack of specialized evaluation datasets and metrics remains a significant challenge. 
This highlights a significant research gap in domain-specific LLM evaluation, particularly in police operations, where responses must align with operational procedures and regulatory requirements. We address this gap by proposing a specialized evaluation framework tailored to this domain.

\section{Framework for Evaluating LLMs in Policing}

Real world policing requires complex judgment and context specific actions, making scenario-based evaluation essential for assessing LLMs. However, no existing research provides a policing specific framework, and relying solely on standard metrics may lead to critical mistakes.

To address this, we propose a \textbf{PAS}, an LLM evaluation framework based on \textbf{Police Action Scenarios}. The \textbf{PAS} is constructed as a five-stage evaluation framework for policing tasks, formally expressed as $E_{\text{police}} = f(S, R, G, M, P)$. Each stage has been designed to ensure consistent applicability across varying police scenarios.

\begin{itemize}
    \item \textbf{Policing Scenario Definition}($S$) : Situation-based task.
    \item \textbf{Reference Responses}($R$) : Construction of golden answers through participation of police experts.
    \item \textbf{Response Generation}($G$) : LLM outputs on scenarios.
    \item \textbf{Core Evaluation Metrics}($M$) : Selection of policing-specific metrics and evaluation methodology design.
    \item \textbf{LLM Performance Evaluation for Policing}($P$) : Comprehensive evaluation of LLM suitability.
\end{itemize}

The \textbf{PAS} framework can be formally represented as:
\begin{equation}
    \text{PAS}: S \xrightarrow{\text{Experts}} R, S \xrightarrow{\text{LLMs}} G \xrightarrow{(M, R)} P
\end{equation}

\subsubsection{\textbf{Step 1: Policing Scenario Definition($S$)}} We proposes a policing scenario-based task design to evaluate LLMs acting in the role of police officers. The objective is to assess the model's reasoning and decision-making in realistic policing contexts. We define scenarios along two key dimensions:

\begin{itemize}
    \item \textbf{Situation (S)}: The operational state in which an officer is placed. This includes the officer's department, role, mission, and the type of incident encountered.
    \item \textbf{Action (A)}: The cognitive or behavioral output required in response to the situation, such as legal judgment, case classification, report generation, or citizen interaction.
\end{itemize}

This formulation is based on core principles. First, legal definitions of police duties, such as crime prevention, investigation, and public order, support the classification of the situation \textbf{\( S \)}~\cite{Korea:judicial-police, Song:police-power}. Policing models, including traditional and intelligence led approaches, inform how officers act across contexts~\cite{pereira2021traditional}, shaping the definition of the action component \textbf{\( A \)}. Since police work relies on legal reasoning and situational judgment~\cite{roberts2012law, clark2012covert}, LLM evaluations must reflect this cognitive aspect.

\subsubsection{\textbf{Step 2: Reference Responses} ($R$)}

To ensure reliable LLM evaluation for Police Action Scenarios, gold-standard reference answers are essential, as no such datasets currently exist in policing. These are built using input from local police experts and training materials, with real case data reviewed and filtered for high-reliability responses. This ensures both jurisdictional relevance and adaptability, providing strong benchmarks and resources for future LLM alignment.

\subsubsection{\textbf{Step 3: LLM Response Generation} ($G$)}
In this stage, LLM responses are generated from the defined Police Action Scenarios, providing empirical insight into how well LLMs mirror police reasoning. These outputs help assess the need for police-specific LLMs. Given the complexity of policing tasks, this step also lays the groundwork for future comparisons using specialized evaluation metrics. Given a \textbf{Police Action Scenario} \( S \), the LLM generates responses \( G = \{ g_1, \dots, g_n \} \) that approximate police reasoning and judgment. Formally, \( G = \text{LLM}(S) \), where the model maps scenarios to outputs under constraints defined by \( S \).

\subsubsection*{\textbf{Step 4: Core Evaluation Metrics} ($M$)}

To evaluate LLMs in policing, proper metrics are essential for assessing response quality. Existing LLM metrics offer useful foundations but must be adapted to the unique demands of police work. For example, policing tasks demand more than accuracy-they require situational fit, legal-procedural alignment, and judgment rationality. Furthermore, questions remain whether conventional metrics alone can fully capture the quality of responses in police-related tasks.

Therefore, this stage defines the systematic process for selecting the final Core Metrics as follows:

\begin{itemize}
    \item First, candidate metrics are drawn from existing LLM evaluation studies and tailored to fit policing scenarios.
    \item Next, the adapted metrics are applied to LLM responses across scenarios, evaluated by both an automated LLM evaluator and human police experts.
    \item Finally, a two-stage filtering process refines the metric set: (1) multiple regression identifies metrics that significantly predict response quality, and (2) correlation analysis validates these, keeping only those positively aligned with expert judgments.
\end{itemize}

\subsubsection*{\textbf{Step 5: LLM Performance Evaluation} ($P$)}

The final stage conducts a comprehensive evaluation using core metrics adapted from existing LLM research to fit policing needs. Key indicators include expert-aligned overall quality, legal and procedural consistency, and real-world applicability. This structured assessment not only validates the LLM’s utility in police scenarios but also informs future development and policy for policing-focused LLM deployment.

The final evaluation integrates metrics into an overall assessment function \( P = h(G, R, M) \), where \( h \) measures the alignment between generated responses \( G \), reference responses \( R \), and metrics \( M \). This produces a multi-dimensional score reflecting LLM performance in policing scenarios, offering both selection benchmarks and diagnostic insights into alignment with policing standards.

\begin{table}[t]
    \centering
    \small % 9pt size, the smallest allowed
    \begin{tabular}{
        >{\raggedright\arraybackslash}p{0.40\columnwidth} % 1열 폭 축소
        >{\raggedright\arraybackslash}p{0.50\columnwidth} % 2열 폭 확대
    }
        \toprule
        % --- 1행: 카테고리 이름 ---
        \textbf{Logical Thinking} & \textbf{Background Knowledge} \\
        \cmidrule(r){1-1} \cmidrule(l){2-2}
        % --- 2행: 세부 지표 목록 ---
        Logical Robustness \newline
        Logical Correctness \newline
        Logical Efficiency &
        Factuality \newline
        Groundness \newline
        Commonsense Understanding \newline
        Numerical Sensitivity \\
        \midrule
        % --- 3행: 카테고리 이름 ---
        \textbf{Problem Handling} & \textbf{User Alignment} \\
        \cmidrule(r){1-1} \cmidrule(l){2-2}
        % --- 4행: 세부 지표 목록 ---
        Comprehension \newline
        Insightfulness \newline
        Completeness \newline
        Metacognition &
        Readability \newline
        Conciseness \newline
        Harmlessness \newline
        Logical Explanation \\
        \bottomrule
    \end{tabular}
    \caption{Candidate Metrics for Model Evaluation. See supplements for definitions.}
    \label{tab:metic}
\end{table}

\section{Experiment}

\subsection{Applying PAS to Real Police Data}

\subsubsection{Simulating Operational Readiness}
In this study, we selected the scenario of \textit{Police Readiness through Operational Reasoning}, which reflects a common form of mental training that police officers engage in before duty. The \textbf{Situation} represents the preparatory phase prior to starting work, and the \textbf{Action} involves mentally simulating responses to anticipated incidents. This practice is rooted in the daily routine of officers who regularly face high-risk situations, enabling them to respond safely and effectively. For example, a patrol officer might think, ``It's the weekend and there could be more domestic violence calls. What's the protocol again?'' Such mental rehearsals often include consulting manuals or past reports to plan appropriate responses in advance.

\begin{table}[tbp]
\centering
\small
\setlength{\tabcolsep}{3pt} % 열 사이의 여백을 3pt로 줄입니다.
\begin{tabular}{llc} % 'Count' 열을 가운데 정렬(c)로 변경
\hline
\textbf{Category} & \textbf{Item} & \textbf{Count} \\
\hline
\multirow{12}{*}{Area}
& Cyber \& Economic Crimes & 17 \\
& Violent \& Serious Crimes & 14 \\
& General Investigation \& Procedure & 14 \\
& Public Safety & 9 \\
& Special Crimes Investigation & 4 \\
& Traffic & 4 \\
& Victim Protection \& Human Rights & 3 \\
& 112 Emergency Calls & 3 \\
& Forensic Science \& Evidence Analysis & 2 \\
& Crime Prevention & 2 \\
& Security & 2 \\
& Police Organization \& Operations & 1 \\
\hline
\multirow{5}{*}{Process}
& Investigation & 51 \\
& Initial Response \& Suppression & 13 \\
& General & 9 \\
& Post-Incident Protection \& Management & 1 \\
& Internal & 1 \\
\hline
\multirow{5}{*}{Type}
& Function-Specific Manuals & 41 \\
& Investigation Checkpoints & 20 \\
& Criminal Law Knowledge & 6 \\
& Investigation Techniques & 4 \\
& Investigation Reporting Guidelines & 4 \\
\hline
\end{tabular}
\caption{Statistics of Police Manuals by Category}
\label{manualstat}
\end{table}

\subsubsection{Reference Answers from Police Manuals} To generate reference answers for the \textit{Police Readiness through Operational Reasoning}, we used 1,602 official police manuals from the Korean National Police Agency. These documents span various domains, including investigation, law enforcement, traffic control, and emergency response, and contain legal references, procedural guidelines, and training materials. Most were in PDF format with 3 to 4 levels of hierarchy, requiring preprocessing for LLM input. We segmented them by section headers and reformatted the content into a standardized \{title, content, question\} structure. With expert guidance, we filtered for quality and compiled 8,348 curated entries. From this set, we created 75 question and answer pairs, as shown in Table~\ref{manualstat}, covering a broad range of police duties and operational contexts. Each pair was constructed by identifying a key question and locating its answer within the same entry, simulating how officers anticipate and reason through situations before beginning their shift.

\subsubsection{LLM Response Generation} We conducted zero shot experiments using the final set of 75 questions with three commercial LLMs: GPT-4, Gemini, and Claude (temperature = 0.8). Each model was given only a brief scenario describing a police officer’s preparatory state before starting duty, and was instructed to respond solely from the perspective of a police officer, outlining the appropriate actions to take without access to any external knowledge or additional context.

\subsubsection{Core Metric Alignment} We then identified the optimal metrics through evaluation by experts, as outlined below.
\begin{itemize}
    \item \textbf{Candidate Metrics Using FLASK}: We began with the 12 core metrics from the FLASK framework~\cite{yeflask}, a validated tool for evaluating LLMs in areas like logical reasoning and problem solving. We expanded this set by splitting factuality into factuality and groundness, and adding numerical sensitivity and logical explanation, inspired by Sun et al.\cite{sun2024principle}. In total, we defined 15 candidate metrics to assess both general LLM abilities and policing-specific needs. Definitions are provided in Table~\ref{tab:metic}.

    \item \textbf{LLM-as-a-Judge \& Police Expert Evaluation}: To identify key metrics for policing-specific applications, we evaluated 225 responses (75 questions across 3 LLMs) and reference answers using the 15 candidate metrics and an overall quality score on a 5 point Likert scale. For the automated assessments, the temperature of the LLM-as-a-Judge was set to 0.8. Both LLM-as-a-Judge and police experts assessed the responses, combining three automated evaluations with two expert reviews. This hybrid method balances prior findings on LLM judge reliability~\cite{hada2023large, chiang2023can} with caution around automating policing-specific evaluation. To reduce this uncertainty, we used structured prompts based on prior LLM evaluation frameworks~\cite{yeflask, zheng2023judging}. Two strategies were employed to enhance reliability. First, each prompt included an expert written reference response alongside the target response, which improves alignment with human judgments~\cite{krumdick2025no} and helps detect domain-specific errors~\cite{ryu2023retrieval}. Second, prompts required judges to explain their reasoning before assigning a score to increase agreement with human evaluators~\cite{chiang2023closer}, especially when guided by a scoring rubric~\cite{pires2025automatic}.

    \item \textbf{Key Metrics Determination}: To identify the most meaningful evaluation metrics, we implemented a two-stage filtering process. First, a stepwise regression method was applied to determine which of the 15 candidate metrics were significant predictors of the overall quality score. Only metrics with a significance level of \textit{p} \textless .05 advanced to the second stage. Next, these selected candidates were validated against the judgments of human police experts using Spearman's rank correlation ($\rho$). A metric was retained for the final set only if it also showed a statistically significant positive correlation with the expert evaluations (\textit{p} \textless .05). This rigorous process was designed to ensure that our final metrics possess both statistical predictive power and domain relevance.
\end{itemize}

\subsection{Experiment Results}

\subsubsection{RQ1: What Is an Appropriate Evaluation Framework for LLMs in Police Operations That Integrates Domain-Specific Metrics and Evaluation Datasets?}
\label{exp:rq1}

\subsubsection{Two-Stage Analysis for Key Metric Selection}
% Identifying Five Key Metrics through Multiple Regression Analysis and Expert Validation

A multiple regression analysis was conducted to determine the most significant metrics for evaluating LLM-generated responses in police operations. Among the 15 candidate metrics, six were found to be statistically significant in predicting response quality. The model exhibited a strong fit ($R^2 = 0.856$, $F = 297.457$, $n = 300$, $p < 0.001$), suggesting that these criteria are essential for the effective evaluation of LLM-generated responses in police operations.

Next, these candidates were validated against the judgments of human police experts using Spearman's rank correlation. As shown in Table~\ref{tab:key-metrics-selection}, five of the six metrics demonstrated a statistically significant positive correlation with the expert evaluations (p \textless .05). However, we note that the correlation coefficients themselves were moderate, ranging from $\rho$=0.315 to $\rho$=0.560, which is consistent with known challenges in automated evaluation of specialized domains~\cite{szymanski2025limitations}. This indicates that while the LLM-as-a-judge scores are directionally aligned with expert opinions, they are not a perfect substitute.

The final key metrics are: Logical Correctness, Completeness, Factuality, Logical Efficiency, and Logical Robustness.

\begin{table}[t]
\centering
\small % 단일 컬럼에 맞추기 위해 폰트 크기를 약간 줄입니다.
\setlength{\tabcolsep}{4pt} % 열 사이의 간격을 약간 줄입니다.
\begin{tabularx}{\columnwidth}{@{}lcccccc@{}}
\toprule
 & \multicolumn{3}{c}{\textbf{Regression}} & \multicolumn{2}{c}{\textbf{Correlation}} & \textbf{Final} \\
\cmidrule(lr){2-4} \cmidrule(lr){5-6}
\textbf{Metric} & \textbf{$\beta$} & \textbf{SE} & \textbf{\textit{p}-val} & \textbf{$\rho$} & \textbf{\textit{p}-val} & \textbf{Sel.} \\ \midrule
L. Correctness & 0.116 & 0.049 & .018 & 0.560 & \textless.001 & $\checkmark$ \\
Completeness & 0.141 & 0.042 & .001 & 0.409 & \textless.001 & $\checkmark$ \\
Factuality & 0.247 & 0.048 & \textless.001 & 0.330 & .002 & $\checkmark$ \\
L. Efficiency & 0.137 & 0.048 & .004 & 0.330 & .002 & $\checkmark$ \\
L. Robustness & 0.167 & 0.052 & .002 & 0.315 & .003 & $\checkmark$ \\ \midrule
L. Explanation & 0.108 & 0.036 & .003 & 0.203 & .055 & $\times$ \\ \bottomrule
\end{tabularx}
\caption{Two-stage analysis for key metric selection. Metrics were selected if they were both statistically significant predictors in multiple regression (\textit{p} \textless .05) and showed a significant positive correlation with expert judgments (\textit{p}
\label{tab:key-metrics-selection}\textless .05).}
\end{table}

\subsubsection{RQ2: To What Extent Can LLMs Effectively Answer Inquiries about Police Work?}

%\subsection{Reference Answer Quality Analysis}
\subsubsection{Reference responses demonstrate high performance across evaluation metrics} The evaluation of the reference answers shows strong overall performance (3.88/5.0), with particularly high scores in factuality (4.15), logical correctness (4.14), and harmlessness (4.11). These results validate the quality of our manual-based reference answers, which serve as the standard for assessing LLMs' performance.

\subsubsection{Identified Key Metrics Reveal Critical Gaps in LLM Performance}

Analysis of the identified key metrics revealed significant deficiencies in LLM performance. As shown in Table~\ref{tab:t-test}, the average performance across Claude, GPT-4, and Gemini was significantly lower than the reference answers across these critical metrics (\( p \textless .05 \)), with an average skill difference of \( -1.115 \) in these key metrics (compared to \( -0.828 \) for all evaluation skills).

\begin{table}[t]
    \centering
    \small
    \setlength{\tabcolsep}{2.5pt} 
    \begin{tabular}{l|D{.}{.}{2}@{\hspace{0.2em}}l|D{.}{.}{2}@{\hspace{0.3em}}l|D{.}{.}{2}@{\hspace{0.3em}}l|D{.}{.}{2}@{\hspace{0.3em}}l} % 간격 미세 조정
        \toprule
        Metric
        & \multicolumn{2}{c|}{All} 
        & \multicolumn{2}{c|}{Claude}
        & \multicolumn{2}{c|}{Gemini}
        & \multicolumn{2}{c}{GPT-4} \\
        \midrule
        \textbf{L. Robustness} & -0.93 & $^{\text{\scriptsize ***}}$ & -0.97 & $^{\text{\scriptsize ***}}$ & -0.72 & $^{\text{\scriptsize ***}}$ & -1.09 & $^{\text{\scriptsize ***}}$ \\
        \textbf{L. Correctness} & \multicolumn{1}{r}{\underline{$-1.32$}} & $^{\text{\scriptsize ***}}$ & \multicolumn{1}{r}{\underline{$-1.34$}} & $^{\text{\scriptsize ***}}$ &  \multicolumn{1}{r}{\underline{$-1.10$}} & $^{\text{\scriptsize ***}}$ & \multicolumn{1}{r}{\underline{$-1.52$}} & $^{\text{\scriptsize ***}}$ \\
        \textbf{L. Efficiency} & \multicolumn{1}{r}{\underline{$-1.09$}} & $^{\text{\scriptsize ***}}$ & -1.08 & $^{\text{\scriptsize ***}}$ & \multicolumn{1}{r}{\underline{$-0.97$}} & $^{\text{\scriptsize ***}}$ & -1.22 & $^{\text{\scriptsize ***}}$ \\
        \textbf{Factuality} & \multicolumn{1}{r}{\underline{$-1.38$}} & $^{\text{\scriptsize ***}}$ & \multicolumn{1}{r}{\underline{$-1.43$}} & $^{\text{\scriptsize ***}}$ & \multicolumn{1}{r}{\underline{$-1.16$}} & $^{\text{\scriptsize ***}}$ & \multicolumn{1}{r}{\underline{$-1.54$}} & $^{\text{\scriptsize ***}}$ \\
        Groundness & \multicolumn{1}{r}{\underline{$-1.41$}} & $^{\text{\scriptsize ***}}$ & \multicolumn{1}{r}{\underline{$-1.48$}} & $^{\text{\scriptsize ***}}$ & \multicolumn{1}{r}{\underline{$-1.17$}} & $^{\text{\scriptsize ***}}$ & \multicolumn{1}{r}{\underline{$-1.57$}} & $^{\text{\scriptsize ***}}$ \\
        Common. Und. & -1.05 & $^{\text{\scriptsize ***}}$ & -1.03 & $^{\text{\scriptsize ***}}$ & -0.86 & $^{\text{\scriptsize ***}}$ & \multicolumn{1}{r}{\underline{$-1.27$}} & $^{\text{\scriptsize ***}}$ \\
        Num. Sensitivity & \multicolumn{1}{r}{\underline{$-1.38$}} & $^{\text{\scriptsize ***}}$ & \multicolumn{1}{r}{\underline{$-1.39$}} & $^{\text{\scriptsize ***}}$ & \multicolumn{1}{r}{\underline{$-1.18$}} & $^{\text{\scriptsize ***}}$ & \multicolumn{1}{r}{\underline{$-1.56$}} & $^{\text{\scriptsize ***}}$ \\
        Comprehension & -1.06 & $^{\text{\scriptsize ***}}$ & -1.11 & $^{\text{\scriptsize ***}}$ & -0.84 & $^{\text{\scriptsize ***}}$ & -1.24 & $^{\text{\scriptsize ***}}$ \\
        Insightfulness & -0.38 & $^{\text{\scriptsize *}}$ & -0.58 & $^{\text{\scriptsize ***}}$ & -0.01 & & -0.54 & $^{\text{\scriptsize ***}}$ \\
        \textbf{Completeness} & -0.86 & $^{\text{\scriptsize ***}}$ & -1.03 & $^{\text{\scriptsize ***}}$ & -0.51 & $^{\text{\scriptsize ***}}$ & -1.04 & $^{\text{\scriptsize ***}}$ \\
        Metacognition & -0.31 & $^{\text{\scriptsize **}}$ & -0.35 & $^{\text{\scriptsize **}}$ & -0.02 & & -0.57 & $^{\text{\scriptsize ***}}$ \\
        Readability & -0.13 & & -0.21 & $^{\text{\scriptsize *}}$ & 0.04 & & -0.23 & \\
        Conciseness & -0.37 & $^{\text{\scriptsize **}}$ & -0.18 & & -0.40 & $^{\text{\scriptsize **}}$ & -0.52 & $^{\text{\scriptsize ***}}$ \\
        Harmlessness & -0.20 & $^{\text{\scriptsize *}}$ & -0.29 & $^{\text{\scriptsize **}}$ & -0.01 & & -0.32 & $^{\text{\scriptsize ***}}$ \\
        L. Explanation & -0.55 & $^{\text{\scriptsize ***}}$ & -0.72 & $^{\text{\scriptsize ***}}$ & -0.24 & & -0.70 & $^{\text{\scriptsize ***}}$ \\
        \midrule
        Overall & -1.02 & $^{\text{\scriptsize ***}}$ & -1.04 & $^{\text{\scriptsize ***}}$ & -0.81 & $^{\text{\scriptsize ***}}$ & -1.19 & $^{\text{\scriptsize ***}}$ \\
        \bottomrule
    \end{tabular}
    \begin{flushleft}
        \footnotesize {$^{*}p < .05$, $^{**}p < .01$, $^{***}p < .001$}
    \end{flushleft}
    \caption{T-test Results on Performance Gaps Between LLMs and Reference Answers. Key metrics identified through multiple regression analysis are in \textbf{bold}, and underlined values represent the five metrics where each LLM shows the greatest deviation from the reference answers, primarily in Logical Thinking and Background Knowledge.}
    \label{tab:t-test}
\end{table}

Model-wise analysis reinforced these findings, with all model-metric combinations showing significantly lower performance than reference answers (\( p < .05 \)), except for Gemini's logical explanation. The performance gap was particularly pronounced in these key metrics. Specifically, Gemini exhibited a difference of \( -0.893 \), Claude \( -1.169 \), and GPT \( -1.283 \), which were notably larger than their respective all-metric average differences of \( -0.611 \), \( -0.880 \), and \( -0.994 \).

This performance gap extends to all evaluation metrics. A comparison with reference answers revealed substantial differences across all models. Specifically, the average difference in overall scores was \( -1.015 \). Statistical analysis using t-tests confirmed that, except for readability, all skills exhibited significantly lower scores compared to reference answers (\( p < .05 \)). These findings reinforce the importance of the identified key metrics and underscore the significant limitations of current LLMs in effectively handling police-work-related queries.

\subsubsection{LLMs are weak in providing high-quality responses to more specialized police knowledge}

\begin{table}[h]
    \centering
    \small
    \setlength{\tabcolsep}{1mm}
    \begin{tabular}{lcccc}
        \toprule
        \textbf{Category} & \textbf{Reference} & \textbf{GPT-4} & \textbf{Gemini} & \textbf{Claude} \\
        \midrule
        Overall Score & 3.88 & 2.69 & 3.06 & 2.83 \\
        Skill Score & 3.72 & 2.73 & 3.11 & 2.84 \\
        \midrule
        Factuality & 4.15 & 2.60 & 2.98 & 2.72 \\
        L. Correctness & 4.14 & 2.61 & 3.04 & 2.80 \\
        Harmlessness & 4.11 & 3.80 & 4.10 & 3.82 \\
        Common. Und. & 4.00 & 2.73 & 3.14 & 2.97 \\
        Groundness & 3.91 & 2.33 & 2.73 & 2.42 \\
        Comprehension & 3.87 & 2.63 & 3.03 & 2.76 \\
        L. Efficiency & 3.84 & 2.63 & 2.87 & 2.77 \\
        Completeness & 3.81 & 2.76 & 3.30 & 2.78 \\
        L. Robustness & 3.81 & 2.72 & 3.09 & 2.84 \\
        Readability & 3.80 & 3.59 & 3.84 & 3.57 \\
        Num. Sensitivity & 3.69 & 2.12 & 2.51 & 2.30 \\
        L. Explanation & 3.60 & 2.90 & 3.37 & 2.89 \\
        Insightfulness & 3.47 & 2.92 & 3.45 & 2.88 \\
        Metacognition & 2.85 & 2.28 & 2.83 & 2.50 \\
        Conciseness & 2.76 & 2.25 & 2.36 & 2.58 \\
        \bottomrule
    \end{tabular}
    \caption{Overall scores of LLMs and comparison of performance across metrics.}
    \label{tab:llm_performance_metrics}
\end{table}

Table~\ref{tab:llm_performance_metrics} compares LLM-generated responses with reference answers, showing that reference answers consistently achieve higher scores in overall performance (3.88) and skill-specific metrics (3.72). As illustrated in Table~\ref{tab:llm_performance_metrics} and~\ref{tab:llm_performance_domains}, this superiority extends across all skill categories and work domains, reinforcing the assumption that police manual-based responses exhibit higher quality than LLM-generated ones. The findings further highlight that LLMs do not consistently produce high-quality responses in all aspects of police work.

A detailed analysis of skill-specific scores indicates that LLMs underperform primarily due to a "lack of expertise in police work." Reference answers score highest in factuality (4.15), logical correctness (4.14), and harmlessness (4.11), demonstrating strong domain-specific knowledge. In contrast, LLMs perform best in harmlessness (3.91), readability (3.67), and insightfulness (3.09), emphasizing their strength in text coherence over factual accuracy.

This gap arises because reference answers originate from police manuals rather than being optimized for LLM prompts. While LLMs generate well-structured text, they lack specificity and factuality in police-related contexts.

Regarding police work areas, answer models handle broad topics but struggle with specialized subjects. Table~\ref{tab:llm_performance_domains} shows that reference answers score highest in Crime Prevention (4.11) and Violent and Major Crimes (4.05). Among answer models, GPT-4 excels in crime prevention (3.33) and victim protection (2.89), while Gemini and Claude achieve their highest scores in police organization and operations. However, all models consistently score low in 112 emergency calls (2.37), traffic (2.55), and security (2.61), highlighting persistent limitations across these domains.

\begin{table}[htbp]
    \centering
    \small
    \setlength{\tabcolsep}{1mm}
    \begin{tabular}{lcccc}
        \toprule
        \textbf{Work Domain} & \textbf{Ref.} & \textbf{GPT-4} & \textbf{Gemini} & \textbf{Claude} \\
        \midrule
        Crime Prevention & 4.11 & 3.22 & 3.33 & 3.11 \\
        Violence \& Major Crimes & 4.05 & 2.86 & 3.21 & 3.02 \\
        Forensic Science & 4.00 & 2.83 & 3.50 & 3.00 \\
        Public Safety & 4.00 & 2.63 & 2.85 & 2.89 \\
        Victim Protection & 4.00 & 2.89 & 3.33 & 3.22 \\
        Traffic & 3.92 & 2.50 & 2.67 & 2.50 \\
        General Investigation & 3.90 & 2.64 & 2.98 & 2.69 \\
        Security & 3.83 & 2.33 & 2.67 & 2.83 \\
        Cyber \& Economic Crimes & 3.78 & 2.63 & 3.10 & 2.98 \\
        112 Report & 3.33 & 2.33 & 2.89 & 1.89 \\
        Special Crime Investigation & 3.33 & 2.67 & 3.25 & 2.33 \\
        \bottomrule
    \end{tabular}
    \caption{Comparison of performance across work domains.}
    \label{tab:llm_performance_domains}
\end{table}

\section{Discussion}

\subsection{PAS Usefulness for LLM Evaluation in Policing}
\begin{figure}[tbp]
    \centering
    \includegraphics[width=0.9\columnwidth]{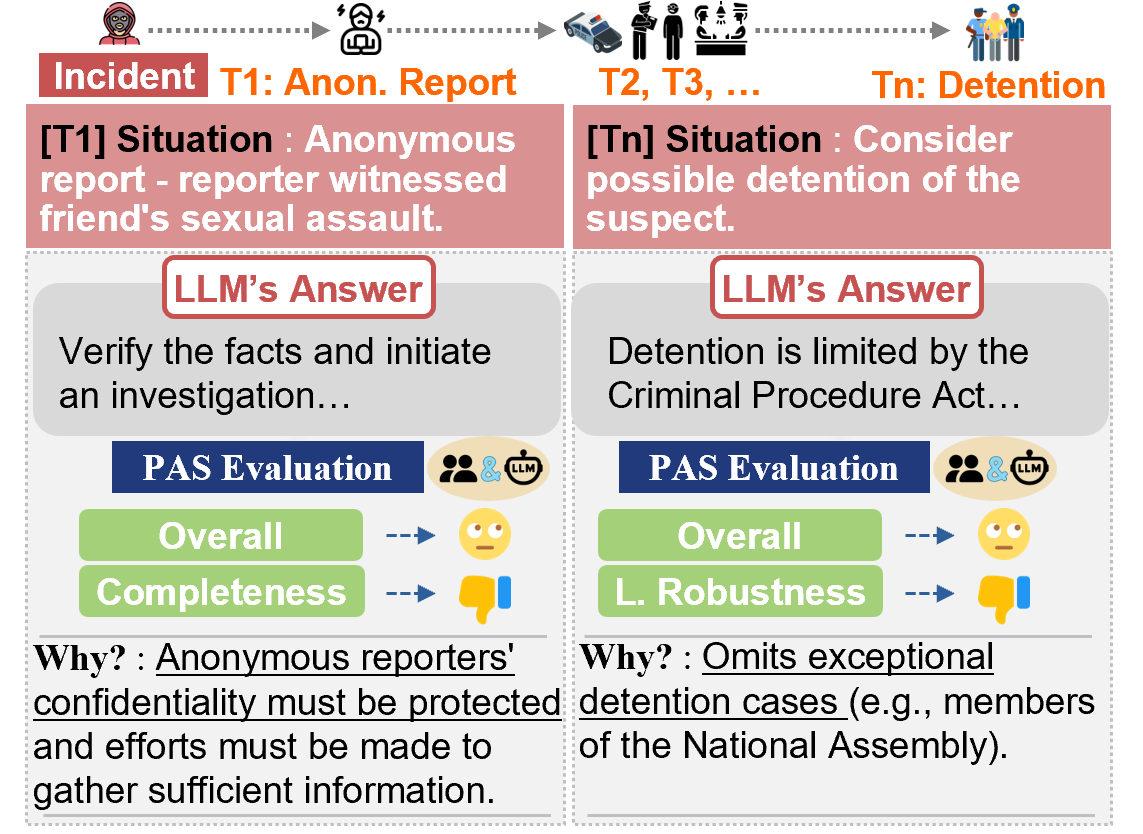}
    \caption{\textbf{PAS} Application in Real Police Work Sequences}
    \label{fig:example}
\end{figure}

We explore how applying the \textbf{PAS} framework enables a more realistic evaluation of LLMs in policing, compared to traditional methods. Figure~\ref{fig:example} presents a case from our experimental scenario, highlighting a typical sequence of tasks drawn from the \textit{Police Readiness through Operational Reasoning}. Although the LLM responses to these questions contain no factual errors and may receive high scores under standard evaluations, they fall short in delivering manual based procedures, practical knowledge, or warnings about legal risks such as unlawful arrest. This suggests that relying on such responses could raise the risk of misconduct. By using \textbf{PAS}, we gain clearer insight into LLM limitations in police contexts and can better identify the outputs that need improvement for practical deployment.

\subsection{Generalizability and Practical Pathways}
A critical challenge for deploying AI in policing is its adaptation to diverse legal and cultural contexts. The PAS framework directly addresses this issue of generalizability through its Reference Responses (R) component. Rather than enforcing a single standard, our framework is designed to be flexibly adapted by having local police experts build the Reference Responses aligned with their own jurisdiction's laws and operational environment. For instance, a reference response based on the Korean Criminal Procedure Act can be replaced with one tailored to U.S. procedures, allowing the same framework structure-with Step 4 enabling metric adaptation as needed for each context-to accurately evaluate LLM performance in a different policing context. This scalable approach enables agencies to measure and improve LLM performance within their unique settings, providing a significant advantage for safe, real-world integration.

\subsection{Implications of Moderate Correlation: The Need for an Expert-in-the-Loop Framework}
The moderate correlation between LLM judges and expert evaluations reveals the limitations of LLM-as-a-judge, reinforcing the importance of our framework's multi-stage validation approach.Recent studies reveal significant limitations of LLM-as-a-judge in expert knowledge domains, with pairwise preference agreement between LLM judges and Subject Matter Experts as low as 60-64\% in specialized healthcare fields~\cite{szymanski2025limitations}. LLMs often align more closely with lay-user preferences than with domain expert criteria~\cite{bavaresco2024llms}, a challenge particularly pronounced in high-stakes policing where training data is scarce and expert judgment is critical. Unlike previous approaches that rely solely on automated metrics, \textbf{PAS} incorporates domain expert validation at key stages, addressing these fundamental limitations of LLM-as-a-judge in specialized fields.

\section{Conclusion}

This study proposes a comprehensive evaluation framework, \textbf{PAS}, developed to address the complexities of police operations. We defined highly realistic police action scenarios and designed a generation method to elicit LLM outputs under those contexts, while also constructing golden answers that enable expert evaluation. As a result, we built QA datasets from over 8,000 police manuals and identified five key indicators—Logical Correctness, Completeness, Factuality, Logical Efficiency, and Logical Robustness—as strong predictors of response quality. Our results demonstrate that commercial LLMs consistently underperform in tasks requiring factual and procedural precision. These findings emphasize the limitations of existing LLMs in specialized domains like policing and the need for research to align outputs with real-world police standards. The \textbf{PAS} not only offers a replicable evaluation benchmark but also provides a practical method applicable to other professional fields. As LLMs become more integrated into public safety systems, such frameworks are vital to ensure reliable and lawful use.

Our experiments also highlight the mixed potential of LLM-as-a-Judge. While scores moderately align with expert assessments, they cannot fully replace human judgment in high-stakes contexts. Moreover, since our evaluation was based on Korean manuals, broader applicability may be limited. Future work should expand \textbf{PAS} with real-time data, jurisdictional diversity, and refined evaluation strategies.

%%%%%%%%%%%%%%%%%%%%%%%%%%%%%%%%%%%%%%%%%%%%%%%%%%%%%%%%%%%%%%%%%%%%%%%%%%%%%%%%%%%%%%%%%%%%%%%%%%%%%%%%%%%%%%%%%%%%%%%%%%%%%%%%%%%%%%%%%%%%%%%%%%%%%%%%%%%%%%%%%%%%%%%%%%%%%%%%%%%%%%%%%%%%%%%%%%%%%%%%%%%%%%%%%%%%%%%%%%%%%%%%%%%%%%%%%%%%%%%%%%%%%%%%%%%%%%%%%%%%%%%%%%%%%%%%%%%%%%%%%%%%%%%%%%%%%%%%%%%%%%%%%%%%%%%%%%%%%%%%%%%%%%%%%%%%%%%%%%%%%%%%%%%%%%%%%%%%%%%%%%%%%%%%%%%%%%%%%%%%%%%%%%%%%%%%%%%%%%%%%%%%%%%%%%%%%%%%%%%%%%%%%%%%%%%%%%%%%%%%%%%%%%%%%%%%%%%%%%%%%%%%%%%%%%%%%%%%%%%%%

\section{Acknowledgments}
This work was supported by the National Research Foundation of Korea(NRF) grant funded by the Korea government(Ministry of Science and ICT) RS-2025-16064585.

\bibliography{aaai2026}

\end{document}